\documentclass{article}
\usepackage{ijcai24}

\usepackage{times}
\usepackage{soul}
\usepackage{url}
\usepackage[hidelinks]{hyperref}
\usepackage[utf8]{inputenc}
\usepackage[small]{caption}
\usepackage{graphicx}
\usepackage{amsmath}
\usepackage{amsthm}
\usepackage{booktabs}
\usepackage{algorithm}
\usepackage{algorithmic}
\usepackage[switch]{lineno}
\usepackage{multirow}
\usepackage{xcolor}

\usepackage{amssymb}
\usepackage[normalem]{ulem}
\useunder{\uline}{\ul}{}
\usepackage[subfigure]{tocloft}
\usepackage{subfigure}
\usepackage{bm}
\usepackage{amsmath}

\title{Make Graph Neural Networks Great Again: A Generic Integration Paradigm of Topology-Free Patterns for Traffic Speed Prediction}

\author{
Yicheng Zhou$^{1,2}$ \and
Pengfei Wang$^{3,4}$ \and
Hao Dong$^{3,4}$\and
Denghui Zhang$^5$\and
Dingqi Yang$^{1,2}$\and \\
Yanjie Fu$^6$\And
Pengyang Wang$^{1,2}$\footnotemark[1]
\affiliations
$^1$The State Key Laboratory of Internet of Things for Smart City, University of Macau, Macau\\
$^2$Department of Computer and Information Science, University of Macau, Macau\\
$^3$Computer Network Information Center, Chinese Academy of Sciences, Beijing\\
$^4$University of Chinese Academy of Sciences, Chinese Academy of Sciences, Beijing\\
$^5$School of Business, Stevens Institute of Technology, Hoboken\\
$^6$School of Computing and AI, Arizona State University, Tempe\\
\emails
\{mc25104, dingqiyang, pywang\}@um.edu.mo,
pfwang@cnic.cn,
donghcn@gmail.com,
dzhang42@stevens.edu,
yanjiefu@asu.edu
}

\begin{document}

\maketitle

\renewcommand{\thefootnote}{\fnsymbol{footnote}}
\footnotetext[1]{Corresponding author.}
\renewcommand{\thefootnote}{\arabic{footnote}}

\begin{abstract}
Urban traffic speed prediction aims to estimate the future traffic speed for improving urban transportation services. 
Enormous efforts have been made to exploit Graph Neural Networks (GNNs) for modeling spatial correlations and temporal dependencies of traffic speed evolving patterns, regularized by graph topology.
While achieving promising results, current traffic speed prediction methods still suffer from ignoring topology-free patterns, which cannot be captured by GNNs. 
To tackle this challenge, we propose a generic model for enabling the current GNN-based methods to preserve topology-free patterns. 
Specifically, we first develop a Dual Cross-Scale Transformer (DCST) architecture, including a Spatial Transformer and a Temporal Transformer, to preserve the cross-scale topology-free patterns and associated dynamics, respectively. 
Then, to further integrate both topology-regularized/-free patterns, we propose a distillation-style learning framework, in which the existing GNN-based methods are considered as the teacher model, and the proposed DCST architecture is considered as the student model. 
The teacher model would inject the learned topology-regularized patterns into the student model for integrating topology-free patterns. 
The extensive experimental results demonstrated the effectiveness of our methods.
\end{abstract}

\section{Introduction}

\begin{figure}[!tbh]
  \centering
  \includegraphics[width=\linewidth]{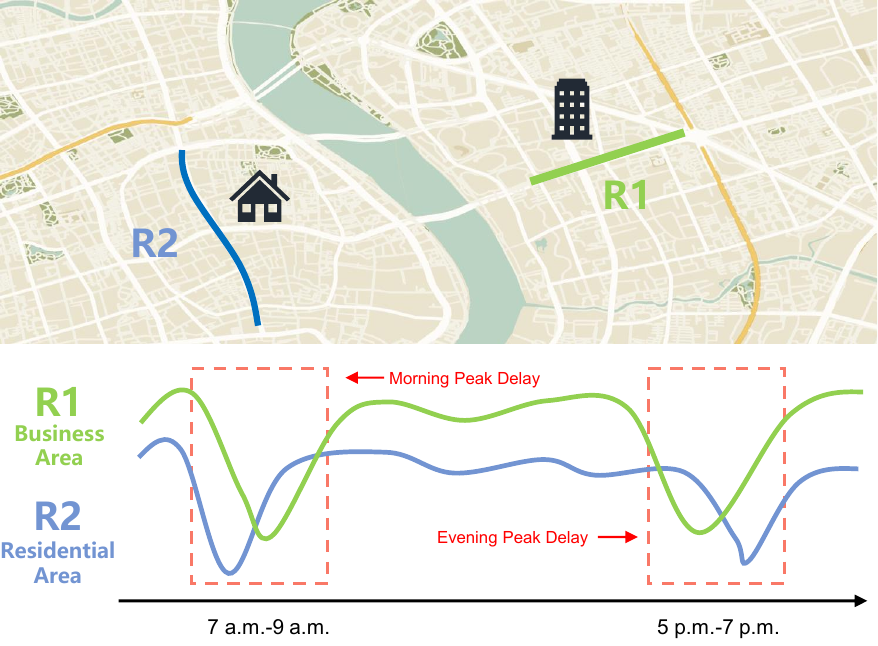}
  \caption{An example of topology-free patterns for traffic speed in the road network, where $R1$ is an arterial road near the business area in the new city district; $R2$ is an arterial road near the residential area in the old city district. During the morning and evening rush hours, the overwhelming traffic on $R1$ and $R2$ causes congestion.}
  \vspace{-1mm}
  \label{figure:motivation}
\end{figure}

Traffic speed prediction, a fundamental task in the development of Intelligent Transportation System (ITS), has gained significant attention in the research community. 
In recent years, methods based on Graph Neural Networks (GNNs) have demonstrated exceptional performance in this area. 
Their effectiveness stems from the capability to model spatial correlations and temporal dependencies through information aggregation across various graph topologies~\cite{dong2023adaptive,hu2023boosting}
, such as road networks~\cite{yu2017spatio,wang2023reinforced}, road segment-segment distance graphs~\cite{li2017diffusion}, variable correlation graphs
~\cite{wu2020connecting,zhang2024spatial,wang2020incremental}, etc.

Despite their promising performance, GNN-based methods are inherently limited by their reliance on topology-regularized patterns. 
During node message propagation, GNNs primarily capture information constrained by the graph topology, leading to the limited scope of data representation. 
This limitation restricts GNNs from recognizing topology-free patterns, referring to latent or indirect relationships beyond the immediate graph structure, which are crucial for traffic speed prediction performances. 
To address the challenge, recent approaches have started to incorporate topology-free patterns.
Notably, attention-based models have been employed to discern the intricate, non-topological relationships between nodes, such as GMAN~\cite{zheng2020gman}, ASTGCN~\cite{guo2019attention}, etc. 

These advancements highlight a trend towards a more sophisticated comprehension of traffic speed dynamics, \textit{combining both topology-regularized and topology-free patterns to improve the predictive accuracy}. 
However, to achieve the goal, three unique challenges arise:

1) \textit{Topology-free patterns vary across different scales.}
Topology-free patterns exhibit different characteristics at different scales.
Figure~\ref{figure:motivation} shows an example of a traffic speed system in the city. 
Note that $R1$ is an arterial road near the business area in the new city district, while $R2$ is an arterial road near the residential area in the old city district. 
At the finest spatial scale, both $R1$ and $R2$ are considered arterial roads in the city, sharing similar functions. 
Consequently, their overall traffic speed patterns exhibit similarity over long periods.
However, when considering a larger spatial scale, the contextual differences of $R1$ and $R2$ remain significant. $R1$, being in the new city district, benefits from more lanes, resulting in an overall higher speed. 
In contrast, $R2$, located in the old city district, has fewer lanes, leading to slower speeds.
This characteristic is not captured at the finest spatial scale. 
Therefore, this phenomenon emphasizes that ``topology-free patterns'' vary across different scales.

2) \textit{Topology-free patterns are dynamically changing.}
Here, we still taking Figure~\ref{figure:motivation} as an example. 
During non-peak hours, the traffic speed patterns of $R1$ and $R2$ are generally similar, since they share similar arterial road characteristics. 
However, when considering $R1$ is located in a business area, while $R2$ is in a residential area, the speed of $R2$ decreases first during the morning peak hours, followed by a decrease in the speed of $R1$. 
This is because residents in the residential area commute to the business area during the morning peak hours. 
When the traffic volume in the residential area increases, it will take several minutes (a delay) to affect the traffic conditions in the business area.
Similarly, during the evening peak hours, people located in the business area will return to the residential area, thus, a similar time delay phenomenon will occur, but the order has changed.
Such difference between $R1$ and $R2$ would occur periodically every day, resulting in short-term divergence of the traffic speed patterns.
Furthermore, as intelligent traffic light controllers are developed to optimize waiting time adaptively\cite{wei2018intellilight}, the dynamics of topology-free patterns become more complicated. 
Therefore, how to capture the dynamics of topology-free patterns is still challenging.

3) \textit{The integration of topology-regularized and topology-free patterns lacks a unified schema.} 
Topology-regularized patterns are typically modeled using GNN-based approaches. 
However, the field of GNNs is characterized by a multitude of variants, each with its unique architecture and method of processing graph-structured data in handling node and edge features~\cite{zhang2020spatio,wang2019adversarial}, varying mechanisms of aggregating neighborhood information~\cite{zhao2019t,wang2018you}, and distinct strategies for capturing the hierarchical and complex patterns within graphs~\cite{wang2019spatiotemporal}.
As a result, each GNN variant offers a different perspective on how to interpret and utilize the topology-regularized patterns in a dataset.
Developing a unified schema that can blend these two types of patterns would enable a more robust and complete analysis of graph-structured data. 
Such an integration is pivotal, especially in complex systems analysis, where both explicit graph structures and implicit, non-structural relationships play crucial roles in shaping the overall dynamics of the system.

Therefore, to tackle the above challenges, we propose a generic framework for boosting current GNN-based traffic speed prediction models by flexibly integrating cross-scale topology-free patterns. 
Specifically, the proposed framework is structured as a two-stage architecture: 
(1) Stage I: Topology-free pattern preservation, where we develop a Dual Cross-Scale Transformer(DCST) by modeling topology-free patterns and dynamics via hierarchical attention interactions across scales in both the spatial and temporal domains;
(2) Stage II: Topology-regularized/-free patterns integration, where we devise a distillation-style integration paradigm that injects topology-regularized into topology-free patterns by regarding the original GNN-base methods as the teacher model and DCST as the student model. 
The proposed integration paradigm is model-agnostic and can serve as a wrapper to apply to any GNN-based model.

In summary, our contributions can be listed as follows:
\begin{itemize}
    \item We identify existing GNN-based methods' limitations and introduce the cross-scale and dynamics of topology-free patterns to the traffic speed prediction task. 
    \item We further propose DCST to effectively capture topology-free patterns.
    \item We devise a distillation-style learning framework to flexibly integrate topology-regularized/-free patterns without bothering to revise the GNN-based models.
    \item We conduct extensive experiments on three real-world datasets to validate the effectiveness of our proposed framework.
\end{itemize}

\begin{figure*}[!t]
	\centering
	\includegraphics[width=1\linewidth]{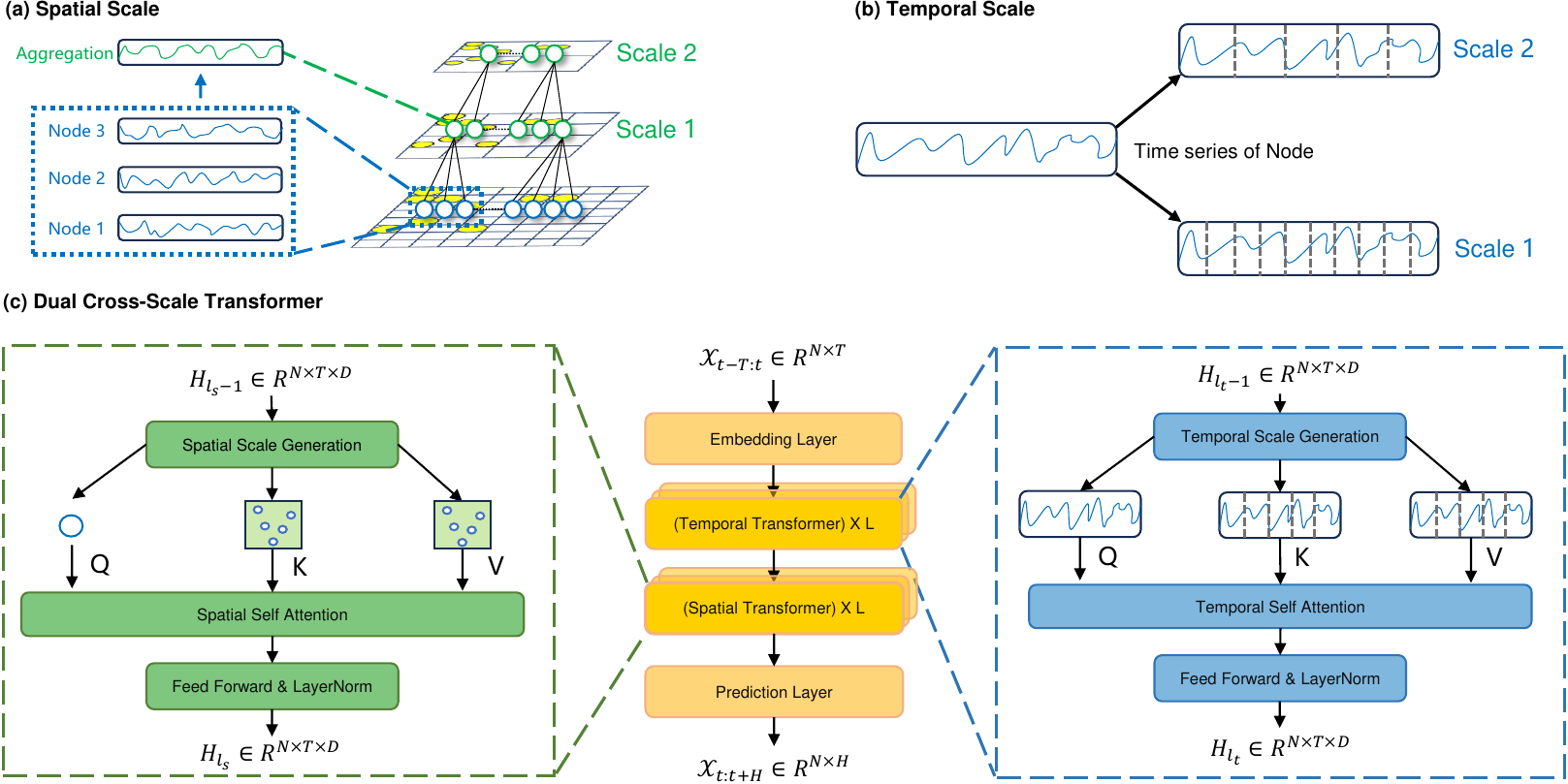}
	\caption{ Framework Overview. 
\textbf{(a)} Spatial Scale: The features of nodes located in the same grid are aggregated, and different scales are divided according to the size of the grid. 
\textbf{(b)} Temporal Scale: For each node, aggregate the features of time points within the same temporal segment and divide them into different scales based on the length of the temporal segment.
\textbf{(c)} Dual Cross-Scale Transformer is composed of an Embedding Layer, a Temporal Transformer, a Spatial Transformer, and a Prediction Layer. }
	\label{fig:Architecture}
\end{figure*}

\section{Problem Formulation}

In this work, we focus on the multi-step traffic speed prediction task that integrates the topology-regularized and topology-free patterns.

Formally, let $\mathcal{G}=(\mathcal{V}, \mathcal{E})$ denote the road network, where $\mathcal{V}=\{v_1, v_2, \dots v_N\}$ represents the road segment set with $N$ segments (recorded by sensors)\footnote{In this work, we interchangeably use road segments, nodes, and sensors.}, and $\mathcal{E}$ is the edge set to demonstrate the adjacency relationship between road segments. 
Each road segment $v_i$ is associated with a $T$-step traffic speed series $\mathbf{x}^{i}=\{ x^{i}_1, x^{i}_2, \cdots, x^{i}_t, \cdots, x^{i}_T \}$, where ${x}^{i}_{t}$ stands for the traffic speed value of $i$-th road segment at $t$-th time step.
Then, the traffic speed records $\mathcal{X}$ of the entire road network $\mathcal{G}$ can be regarded as a multi-variate time series: $\mathcal{X} = \{ \mathbf{x}^{1}, \mathbf{x}^{2}, \cdots, \mathbf{x}^{N}\} \in \mathbb{R}^{N \times T}$.

Following the classic setting in auto-regressive time series forecasting, given the historical observations $ \mathcal{X}_{t-T:t}=\{ \mathbf{x}^{1}_{t-T:t}, \mathbf{x}^{2}_{t-T:t}, \cdots,  \mathbf{x}^{N}_{t-T:t}  \}$ of a certain time period $T$, we aim to predict future traffic speed in a period of time period $H$, denoted by $\mathcal{X}_{t:t+H}=\{\mathbf{x}^{1}_{t:t+H}, \mathbf{x}^{2}_{t:t+H}, \cdots, \mathbf{x}^{N}_{t:t+H}\}$. Then, the traffic speed prediction problem with the integration of topology-regularized and topology-free patterns can be formulated as: 
\begin{equation}
  \mathcal{X}_{t:t+H}  = f(g_1(\mathcal{X}_{t-T:t}, \mathcal{G}), g_2(\mathcal{X}_{t-T:t}))
\end{equation}
where $g_1$ is a learnable function to capture the \textit{topology-regularized patterns} by considering the road network topology, $g_2$ is a learnable function to automatically preserve the \textit{topology-free patterns} without any prior geographical knowledge, and $f$ is a learnable integration function for prediction.

Noted that since we aim to provide a flexible and generic framework for boosting current GNN-based methods, and current GNN-based methods have already been working well in capturing topology-regularized patterns, we directly adopt the current GNN-based methods as $g_1$, and study how to design the topology-free patterns function $g_2$ and the integration function $f$. 
Moreover, the current GNN-based methods ($g_1$) have inherently captured the temporal dependencies~\cite{dong2024temporal}, we will not additionally introduce how to model the temporal dependencies to avoid redundancy.
\section{Methodology}
In this section, we introduce our proposed framework for boosting traffic speed prediction tasks. 
We start with an overview and present each component in detail.

\subsection{Framework Overview} 
\begin{figure}
  \centering
  \includegraphics[width=1\linewidth]{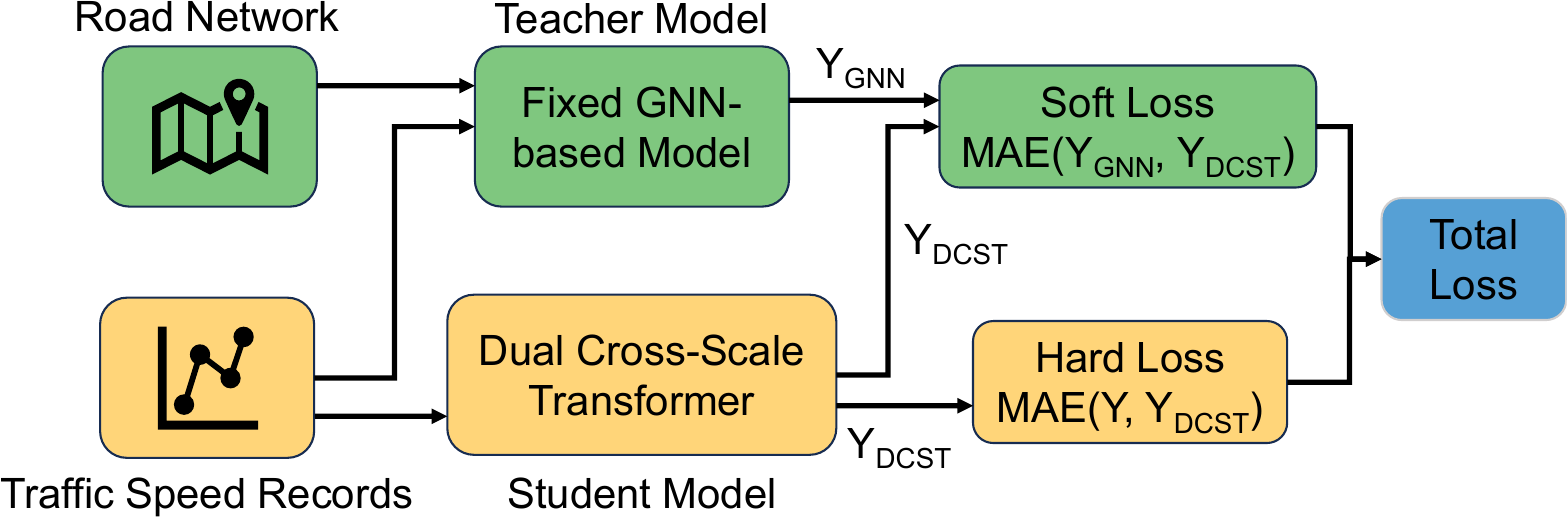}
  \caption{An illustration of the integration process of topology-regularized/-free patterns. The integration process follows the teacher-student paradigm, where the GNN-based model is taken as the teacher model (topology-regularized patterns), and the Dual Cross-Scale Transformer is taken as the student model (topology-free patterns). During the process, the GNN-based model has been pre-trained and kept fixed. The integration is conducted by jointly optimizing ``Soft Loss" and ``Hard Loss".}
  \vspace{-2mm}
  \label{figure:kd}
\end{figure}

Our framework aims to provide a generic wrapper-style solution to enhance the current GNN-based methods by integrating cross-scale topology-free patterns. 
The proposed framework includes two stages: (1) preserving topology-free patterns, and (2) integrating topology-regularized/-free patterns. 
Specifically, in Stage I, we design a Dual Cross-Scale Transformer(DCST) to capture cross-scale topology-free patterns and corresponding dynamics (as shown in Figure~\ref{fig:Architecture}). 
In Stage II, a teacher-student learning framework (as shown in Figure~\ref{figure:kd}) is proposed to integrate topology-regularized/-free patterns, in which the current GNN-based methods are taken as the teacher model, and DCST as the student model. 
The learning framework extracts the knowledge of the topology-regularized spatial correlations from the current GNN-based methods and then passes it into the DCST for integration. 
Then, the well-trained DCST learned through the teacher-student framework will generate predictions by taking into account both the topology-regularized and topology-free patterns. 
In the following content, we will introduce the DCST and the integration procedure in detail.

\subsection{Dual Cross-Scale Transformer for Topology-Free Patterns}
Transformer networks~\cite{vaswani2017attention} have emerged as a predominant paradigm in the realms of natural language processing and computer vision. 
The core idea of Transformer is to exploit the self-attention mechanism to automatically explore the correlations and dependencies among the input tokens. 
As discussed, the topology-free patterns indicate the complex interactions among nodes that are beyond the graph topology modeled by GNN-based methods. 
Therefore, we develop a new Dual Cross-Scale Transformer  (DCST) to preserve topology-free patterns and corresponding dynamics.

Next, we introduce how to divide data into different scales at the spatial and temporal dimensions. 
And then, we present the design of the proposed DCST.

\subsubsection{Spatial-Temporal Scale Generation}

\paragraph{Spatial Scale.} 
We split the geospace into grids based on a pre-defined standard ({\it i.e.}, width and length). 
Different standards lead to different scales. 
Nodes (road segments) are distributed in the grids. 
Let $\mathbf{\mathbf{h}}^i_{t-T:t} \in \mathbb{R}^{T \times D}$ denote the $D$-dimensional representation of the $i$-th node, then the representation matrix can be represented as 
\begin{equation}
    \mathbf{H}_{t-T:t}=[\mathbf{\mathbf{h}}^1_{t-T:t}, \cdots, \mathbf{\mathbf{h}}^i_{t-T:t}, \cdots, \mathbf{\mathbf{h}}^N_{t-T:t}].
\end{equation}
where $\mathbf{H}_{t-T:t} \in \mathbb{R}^{N \times T \times D}$. 
Then, we represent the grid representation by aggregating the associated nodes. 
Formally, given the $m$-th grid of the $l_s$-th spatial scale, the representation $\mathbf{Z}^{m}_{l_s}$ can be represented as 
\begin{equation}
    \mathbf{Z}_{l_s}^{m} = \textmd{LN} \Bigl( \sum \limits_{\forall i\in \Gamma_{l_s}(m)} \  (\mathbf{h}_{t-T:t}^{i} \mathbf{W}^i_{l_s}  + \mathbf{b}^i_{l_s}) \Bigr), 
\end{equation}
where $\mathbf{Z}_{l_s}^{m} \in \mathbb{R}^{T \times D}$, and $\text{LN}$ denotes layer normalization, and   
$ \mathbf{W}^i_{l_s} \in {\mathbb{R}}^{D \times D}$ and $ \mathbf{b}^i_{l_s} \in {\mathbb{R}}^{D}$ denote the weight and bias terms, respectively. 
Then, we denote the representation matrix for the $l_s$-th spatial scale as $\mathbf{Z}_{l_s}$.

\paragraph{Temporal Scale.} 
To capture the dynamics of topology-free patterns, we construct temporal scales by splitting the observations in terms of different unit time lengths. 
The larger the length, the coarser the scale. 
Let $\xi_{l_t}$ denote the unit time length for the $l_t$-th temporal scale, then the constructed temporal scales can be represented as
\begin{equation}
\begin{aligned}
     \mathbf{h}^{i}_{t-T:t} &= \{\mathbf{S}_{j,l_t}^{i} \mid 1 \le j \le \frac{T}{\xi_{l_t}}, \enspace 1\le i \le N \}, \\
    \mathbf{S}_{j,l_t}^{i} &= \{\mathbf{h}_{t}^{i} \mid (j-1) \times \xi_{l_t}  < t \le j \times \xi_{l_t} \},
\end{aligned}
\label{eq:temporal_scale}
\end{equation}
where ${\mathbf{S}}_{j,l}^{i}  \in  {\mathbb{R}}^{{\xi_{l_t} \times D}}$ is the $j$-th segment of node $v_i$ on the $l_t$-th temporal scale. 
For convenience, we set $\xi_{l_t}$ divisible by $T$. 
Then, the representation of the $j$-th segment of node $v_i$ on the $l_t$-th temporal scale $\mathbf{P}_{j,l_t}^{i}$ can be represented as
\begin{equation}
    \mathbf{P}_{j,l_t}^{i} = \textmd{LN}  (  \mathbf{S}_{j,l_t}^{i} \mathbf{W}_{j,l_t}  + \mathbf{b}_{j,l_t}  ), 
\label{eq:temporal_scale_weight}
\end{equation}
where $\mathbf{P}_{j,l_t}^{i} \in {\mathbb{R}}^D$, $\mathbf{S}_{j,l_t}^{i}$ is reshaped as $\mathbb{R}^{1 \times (D \times \xi_{l_t})}$, and  
$\mathbf{W}_{j,l_t} \in {\mathbb{R}}^{(D\times \xi_{l_t}) \times D}$ and $\mathbf{b}_{j,l_t} \in \mathbb{R}^D$ denote weight and bias terms, respectively. 
Then, the representation matrix of the $l_t$-th temporal scale can be denoted as $\mathbf{P}_{l_t}$. 
To reduce the complexity, all nodes at the same segment of a given temporal scale share the same parameters.

By adjusting the unit length of $\xi_{l_t}$, we can obtain temporal segments with different scales, which are used to capture the dynamics of cross-scale topology-free patterns.

\subsubsection{Dual Cross-Scale Transformer} 
The proposed Dual Cross-Scale Transformer (DCST) is composed of an embedding layer (denoted as $\text{FC}$), a temporal Transformer, a spatial Transformer, and a prediction layer. 
Specifically, the embedding layer is a fully connected layer that transforms the original data into $D$-dimensional representations, represented as 
\begin{equation}
    \mathbf{H}_{t-T:t} = \textmd{FC}  (  \mathcal{X}_{t-T:t} ).
\end{equation}

Then, the Temporal Transformer takes the representation $\mathbf{H}_{t-T:t}$ as input to capture the dynamics of the topology-free patterns by investigating attention between nodes across temporal segments in different scales. 
Then, the updated node representations are further fed into the Spatial Transformer to capture the cross-scale characteristics of the topology-free patterns by exploring the relationship between nodes and affiliated grids. 
The Spatial and Temporal Transformers follow the same architecture. 
Each layer of the Transformers corresponds to one spatial or temporal scale, respectively.
For a general description, we utilize ``$\ast$'' to represent $s$ (spatial) or $t$ (temporal), and ignore the time subscript ``$t-T:t$'' to denote the learned node representation by the $l_\ast$-th layer as $\mathbf{H}_{l_\ast}$. 
Then, the updating process of the node representation on the $l_\ast$-th layer can be represented as 
\begin{equation}
\begin{aligned}
    \widetilde{\mathbf{H}}_{l_\ast} &= \textmd{LN} \Bigl( \mathbf{H}_{l_\ast-1}+ \textmd{MSA}(\mathbf{\mathbf{H}}_{l_\ast-1} \mathbf{W}^Q_{l_\ast}, \mathbf{\Phi}_{l_\ast} \mathbf{W}^{K}_{l_t\ast},\mathbf{\Phi}_{l_\ast} \mathbf{W}^{V}_{l_\ast}) \Bigr), \\
    \mathbf{H}_{l_\ast} &= \textmd{LN} \Bigl( \widetilde{\mathbf{H}}_{l_\ast} + \textmd{MLP}(\widetilde{\mathbf{H}}_{l_\ast}) \Bigr) ,
\end{aligned}
\label{eq:cross_scale}
\end{equation}
where $\mathbf{\Phi}$ denotes $\mathbf{Z}_{l_s}$ for the Spatial Transformer, and $\mathbf{P}_{l_t}$ for the Temporal Transformer, respectively; 
$ \mathbf{W}^{Q}_{l_\ast}$, $\mathbf{W}^{K}_{l_\ast}$, and $\mathbf{W}^{V}_{l_\ast}$ are learnable parameters; $\textmd{MSA}(Q, K, V)$ is multi-head self-attention block and $Q,K,V$ serves as queries, keys and values~\cite{vaswani2017attention}; 
$\textmd{MLP}$ represents a multi-layer feedforward block~\cite{liu2021pyraformer}. 
Suppose there are $L_s$ spatial scales and $L_t$ temporal scales, then, the output of the Temporal Transformer can be denoted as $\mathbf{H}_{L_t}$. 
The Spatial Transformer takes $\mathbf{H}_{L_t}$ as input, and the output can be denoted as $\mathbf{H}_{L_s}$.
In the process of aggregating cross-scale topology-free patterns, DCST aggregates $\mathbf{\Phi}$ features from fine-grained to coarse-grained.
This means that in the Temporal Transformer, the length of $\xi_{l_t}$ becomes longer as the number of layers increases, and in the Spatial Transformer the size of the grid changes from small to large.

The prediction layer is a fully connected layer, which takes the learned representation $\mathbf{H}_{L_s}$ as input and then generates prediction $Y$.
\begin{equation}
    Y = FC(\mathbf{H}_{L_s}).
\end{equation}

\begin{table*}[!th]
\centering
\small

\setlength\tabcolsep{2.8mm}

\begin{tabular}{c|ccc|ccc|ccc}

\toprule
\multirow{2.7}{*}{Methods} & \multicolumn{3}{c|}{METRLA} & \multicolumn{3}{c|}{PEMSBAY} & \multicolumn{3}{c}{PEMSD7(M)} \\ \cmidrule{2-10}
 & MAE & RMSE & MAPE & MAE & RMSE & MAPE & MAE & RMSE & MAPE \\ \midrule
HA & 11.01 & 14.74 & 23.34\% & 3.33 & 6.69 & 8.10\% & 3.92 & 7.08 & 9.92\% \\
LSTNet & 4.89 & 9.74 & 11.74\% & 2.26 & 4.23 & 4.94\% & 3.10 & 5.50 & 7.51\% \\ 
\midrule
GMAN & 4.46 & 10.11 & 12.01\% & 1.88 & 4.35 & 4.42\% & 3.22 & 6.48 & 8.19\% \\
ASTGCN & 4.46 & 9.62 & 11.45\% & 1.75 & 4.28 & 4.04\% & 2.85 & 5.44 & 7.40\% \\
STAEformer & 2.94 & 5.98 & 8.10\% & 1.56 & 3.46 & 3.50\% & 2.56 & 5.17 & 6.42\% \\ 
STID & 3.12 & 6.51 & 9.14\% & 1.56 & 3.60 & 3.50\% & 3.04 & 6.25 & 7.73\% \\
\midrule
\midrule
STGCN & 3.67 & 6.51 & 10.21\% & 2.28 & 4.21 & 5.08\% & 3.97 & 6.72 & 9.95\% \\
STGCN - KD & 3.00 & 5.99 & 8.31\% & 1.59 & 3.55 & 3.56\% & 2.60 & 5.13 & 6.57\% \\
\textit{Improvement} & +18.26\% & +7.99 & +18.61\% & +30.26\% & +15.68\% & +29.92\% & +34.51\% & +23.66\% & +33.97\% \\
\midrule
DCRNN & 3.13 & 6.28 & 8.64\% & 1.68 & 3.75 & 3.84\% & 2.88 & 5.88 & 7.10\% \\
DCRNN- KD & 2.97 & 5.94 & 8.15\% & \textbf{1.55} & 3.50 & 3.49\% & 2.56 & 5.11 & 6.47\% \\
\textit{Improvement} & +5.11\% & +5.41\% & +5.67\% & +7.74\% & +6.67\% & +9.11\% & +11.11\% & +13.10\% & +8.87\% \\
\midrule
GWNet & 3.05 & 6.04 & 8.47\% & 1.60 & 3.57 & 3.61\% & 2.59 & 5.07 & 6.46\% \\
GWNet - KD & \textbf{2.92} & \textbf{5.87} & 8.07\% & 1.56 & \textbf{3.45} & 3.48\% & \textbf{2.52} & \textbf{5.01} & \textbf{6.34\%} \\
\textit{Improvement} & +4.26\% & +2.81\% & +4.72\% & +2.50\% & +3.36\% & +3.60\% & +2.70\% & +1.18\% & +1.86\% \\
\midrule
MTGNN & 3.08 & 6.23 & 8.30\% & 1.59 & 3.55 & 3.54\% & 2.62 & 5.16 & 6.44\% \\
MTGNN - KD & 2.98 & 5.98 & \textbf{8.04\%} & 1.56 & 3.49 & \textbf{3.45\%} & 2.53 & 5.07 & 6.37\% \\
\textit{Improvement} & +3.25\% & +4.01\% & +3.13\% & +1.89\% & +1.69\% & +2.54\% & +3.44\% & +1.74\% & +1.09\% \\
\midrule
AGCRN & 3.17 & 6.33 & 8.85\% & 1.64 & 3.66 & 3.70\% & 2.64 & 5.33 & 6.57\% \\
AGCRN - KD & 3.01 & 6.06 & 8.36\% & 1.58 & 3.52 & 3.53\% & 2.56 & 5.06 & 6.39\% \\
\textit{Improvement} & +5.05\% & +4.27\% & +5.54\% & +3.66\% & +3.83\% & +4.59\% & +3.03\% & +5.07\% & +2.74\% \\

\bottomrule
\end{tabular}

\caption { Overall performance comparison. The ``Improvement'' indicates the performance improvement of GNN-based model when combined with DCST. The best performances are highlighted in \textbf{bold} fonts. }
\vspace{-2mm}

\label{table:Overall_comparison}

\end{table*}

\subsection{Integration of Topology-regularized and Cross-Scale Topology-free Patterns}

The proposed Dual Cross-Scale Transformer has deeply explored the cross-scale topology-free patterns in traffic speed prediction tasks. 
However, we still have another research question: can we further integrate the topology-regularized patterns with the topology-free patterns to boost traffic speed prediction performances of GNN-based methods? 
To this end, we propose a novel teacher-student framework to conduct the integration through knowledge distillation. 
Specifically, we take the current GNN-based methods as the teacher model and the proposed DCST as the Student model. 
Intuitively, since the GNN-based methods are graph-based models that leverage the topology of the graph to describe the relationships between sensors and GNNs to capture correlations, they can effectively capture topology-regularized patterns. 
Through knowledge distillation, the topology-regularized patterns are learned and then passed to the DCST for integration. 
Formally, let $Y_{\text{GNN}}$, $Y_{\text{DCST}}$, and $Y$ denote the predictions of the GNN-based methods (the teacher model), DCST (the student model), and the ground truth of the traffic speed. 
We first pre-train the GNN-based methods to fit the ground truth. 
Then, we fix the GNN-based methods and conduct the integration process by optimizing DCST with the help of GNN-based methods. 
Specifically, the integration has two objectives: (1) accepting the knowledge from the GNN-based methods, and (2) predicting as accurately as possible. 
Therefore, following the convention of the teacher-student paradigm, the training loss can be represented as: 
\begin{equation}
    \mathcal{L} = \alpha \cdot \underbrace{\text{MAE}(Y_{\text{DCST}},Y_{\text{GNN}})}_{\text{Soft Loss}} + \beta \cdot \underbrace{\text{MAE}(Y_{\text{DCST}},Y)}_{\text{Hard Loss}} 
    \label{equation:kd}
\end{equation}
where $\text{MAE}$ denotes Mean Square Error, and $\alpha$ and $\beta$ are hyperparameters for Soft Loss and Hard Loss, respectively. 
Specifically, the ``Soft Loss'' is to set the prediction results of the GNN-based methods $Y_{\text{GNN}}$ as the target, and push the prediction of DCST $Y_\text{DCST}$ as close as to the GNN-based methods. 
Along this line, the learned topology-regularized patterns will be integrated into the DCST. 
On the other hand, the ``Hard Loss'' aims to make the DCST generate precise prediction results, which can provide the correct optimization direction for the integration. 
The integration is conducted automatically by minimizing $\mathcal{L}$ in Equation~(\ref{equation:kd}). 

\section{Experiment}

In this work, we conduct extensive experiments on three real-world datasets to evaluate the performance of our proposed methods in traffic prediction tasks. Particularly, our experiments aim to answer the following research questions:
\begin{itemize}
    \item \textbf{RQ1:} How well does our framework perform in traffic prediction tasks? Can our proposed framework boost current GNN-based methods?
    \item \textbf{RQ2:} How much can topology-regularized and cross-scale topology-free patterns contribute to traffic speed prediction, respectively? 
    \item \textbf{RQ3:} How do the key components of the Dual Cross-Scale Transformer architecture contribute to the results? 
\end{itemize}

\begin{figure*}[!htbp]
    \centering
    \subfigure[\scriptsize{METRLA}]{
        \includegraphics[width=0.3\linewidth]{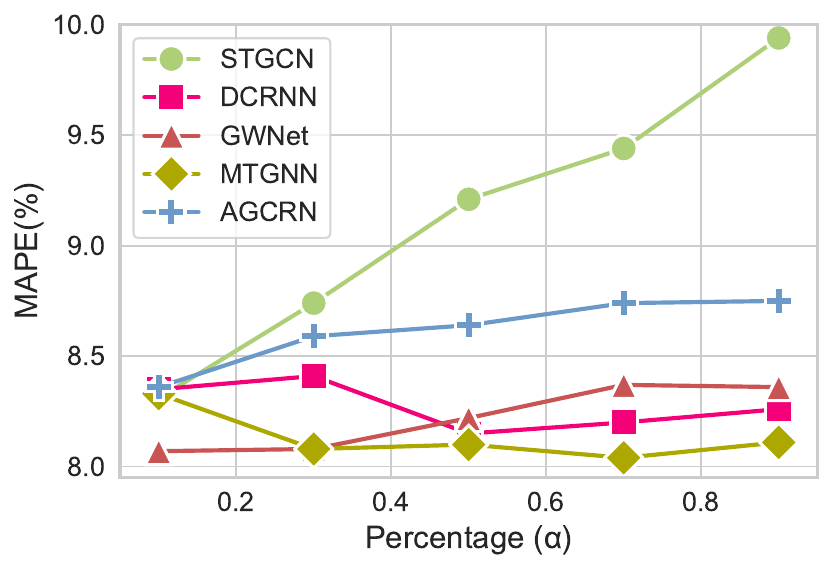}
        }
        \hspace{2mm}
        \vspace{-1mm}
    \subfigure[\scriptsize{PEMSBAY}]{
        \includegraphics[width=0.3\linewidth]{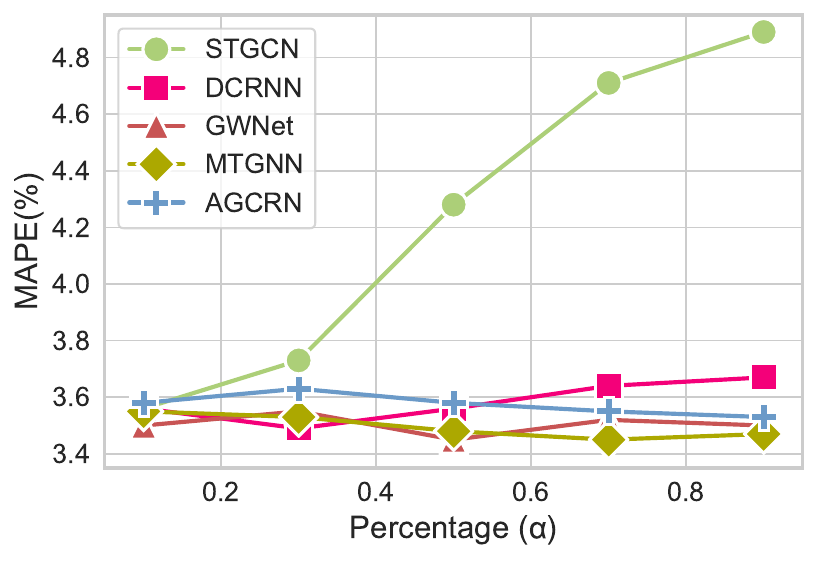}
        }
        \hspace{2mm}
        \vspace{-1mm}
    \subfigure[\scriptsize{PEMSD7(M)}]{
        \includegraphics[width=0.3\linewidth]{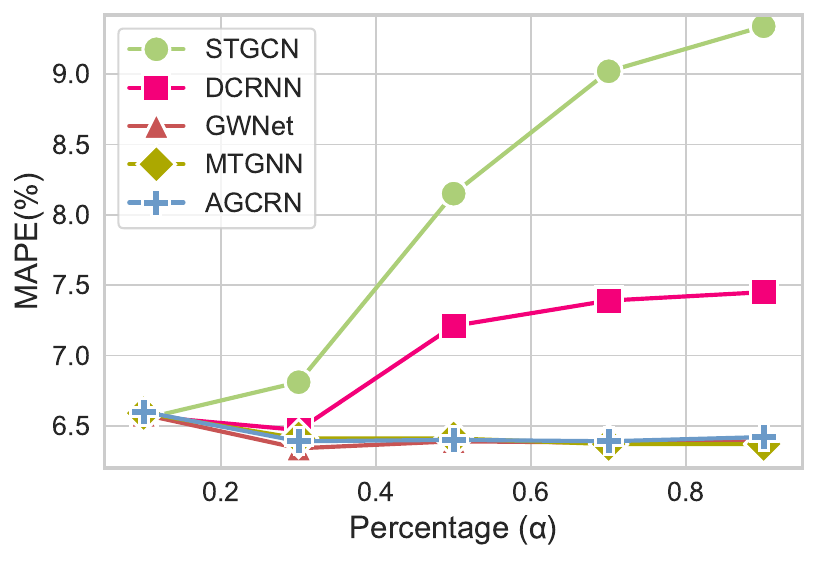}
        }
        \hspace{2mm}
        \vspace{-1mm}
    \vspace{1mm}
    \caption{An illustration of DCST performances w.r.t. different trade-off parameter pairs. We present the results of MAPE on METRLA, PEMSBAY and PEMSD7(M).}
    \vspace{-3mm}
    \label{fig:ex_im}
\end{figure*}

\vspace{-1.5mm}
\subsection{Experiment Setting}

\noindent\textbf{Datasets.}
We evaluate our proposed framework on three traffic speed datasets~\cite{jiang2021dl}, including: ``METR-LA''(Los Angeles), ``PEMS-BAY''(San Francisco), and ``PEMSD7(M)''(California). 
In the experiment,  we split the datasets into three non-overlapping sets, where the earliest $70\%$ of the data is the training set, the following $20\%$ are validation set, and the remaining $10\%$ of the data are test set. Our implementation is available in Pytorch3\footnote{\url{https://github.com/ibizatomorrow/DCST}}.

\vspace{0.5mm}

\noindent\textbf{Comparsion Setup.}
Since our proposed framework is a generic wrapper for boosting current GNN-based models, we evaluate the performance following an ablation study manner. 
Specifically, we take five widely-used GNN-based models as the base models and compare their performance with/without our proposed framework. 
The selected five base models are STGCN~\cite{yu2017spatio}, DCRNN\cite{li2017diffusion}, GWNet\cite{wu2019graph}, MTGNN\cite{wu2020connecting} and AGCRN\cite{bai2020adaptive}. 
We also compare non-GNN-based models for a broader analysis, including two linear models: HA and LSTNet\cite{lai2018modeling}, and four models that solely capture topology-free patterns: GMAN, ASTGCN, STAEformer (based on attention)~\cite{liu2023spatio}, and STID (based on MLP)~\cite{shao2022spatial}.

When applying our proposed framework, we take the GNN-based model as the teacher model, and the proposed DCST as the student model. 
We denote the GNN-based model powered by our framework as ``$\ast$-KD'', where $\ast$ refers to the GNN-based model, such as STGCN-KD, DCRNN-KD, GWNet-KD, MTGNN-KD, and AGCRN-KD, respectively.

\vspace{0.5mm}

\noindent\textbf{Evaluation Metrics.}
We select three widely used metrics for traffic speed prediction tasks, including  Mean Absolute Error (MAE), Root Mean Squared Error (RMSE), and Mean Absolute Percentage Error (MAPE). 
We use historical 12 time steps to predict future average 12 time steps.

\vspace{-1mm}
\subsection{RQ1: Overall Comparison}

Table~\ref{table:Overall_comparison} summarizes the overall experimental results. The bold results are the best. 
Based on Table~\ref{table:Overall_comparison}, we can make the following observations and analysis: 
(1) When our framework combines different GNN-based models, the performance can exceed models that only capture topology-free patterns. 
This reflects the importance of considering both topology-regularized and cross-scale topology-free patterns.
(2) All the enhanced “*-KD” versions consistently outperform the base versions in terms of all metrics over all datasets, these results can validate our motivation that the cross-scale topology-free patterns are essential for boosting traffic speed prediction. 
Such wrapper-style design benefits the current SOTA GNN-based models without modifying the original architecture but only needs to pass the learned topology-regularized patterns for integration.

\subsection{RQ2: Analysis of Topology Regularized/-Free Patterns}

\begin{figure*}[!ht]
    \centering
    \subfigure[\scriptsize{MAE}]{
        \includegraphics[width=0.32\linewidth]{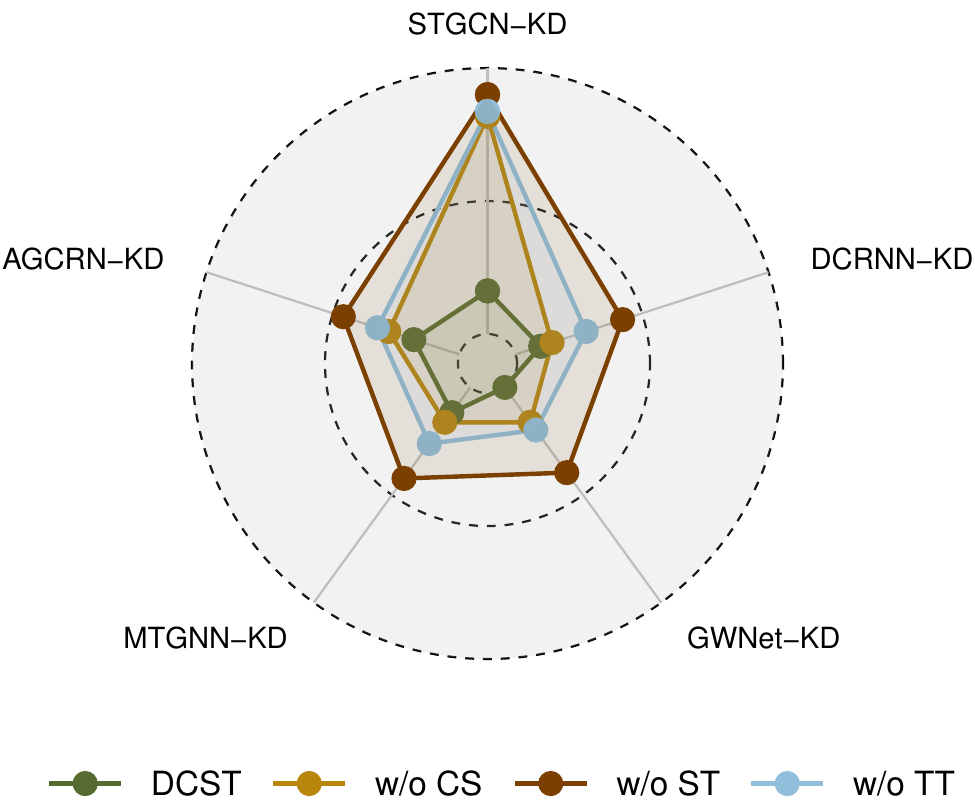} 
        }
        \hspace{-2mm}
        \vspace{0mm}
    \subfigure[\scriptsize{RMSE}]{
        \includegraphics[width=0.32\linewidth]{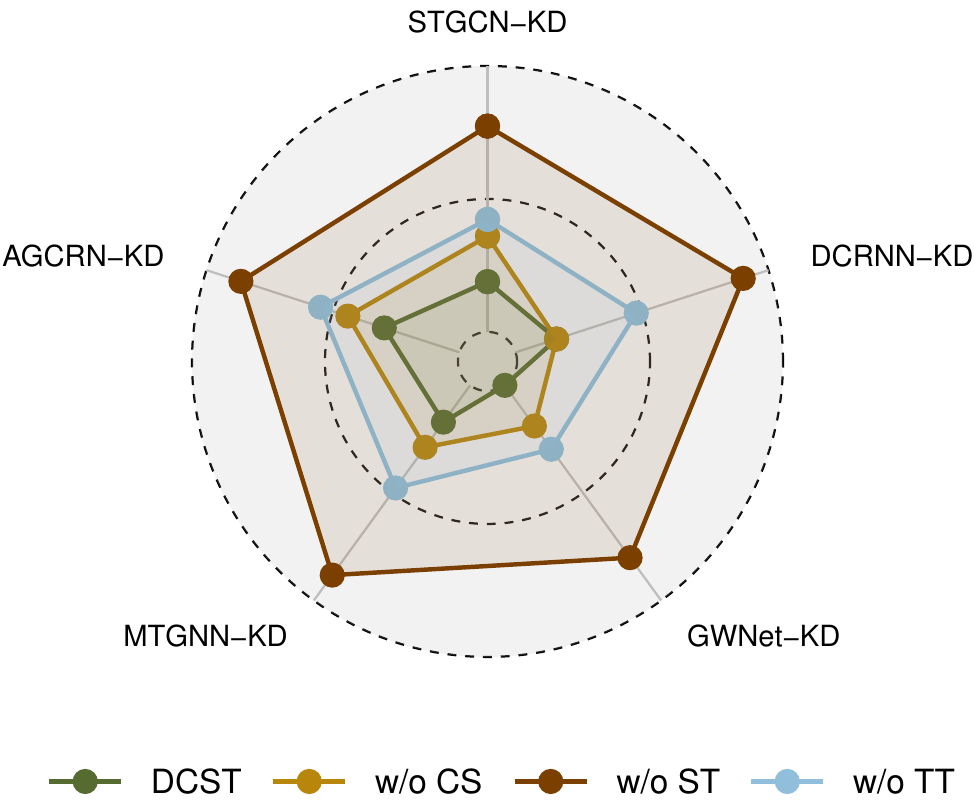}
        }
        \hspace{-2mm}
        \vspace{0mm}
    \subfigure[\scriptsize{MAPE}]{
        \includegraphics[width=0.32\linewidth]{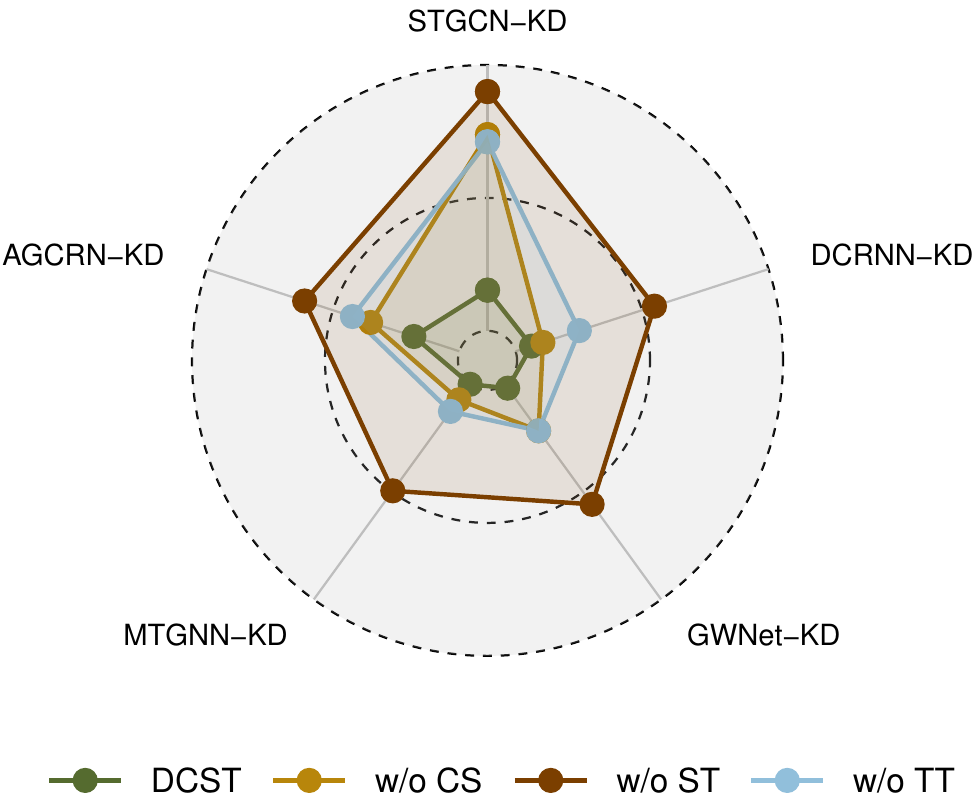}
        }
        \hspace{-2mm}
        \vspace{0mm}
    \caption{Ablation Studies of STGCN-KD, DCRNN-KD, GWNet-KD, MTGNN-KD and AGCRN-KD on metrics MAE, RMSE and MAPE on the METRLA dataset.}
    \vspace{-2.5mm}
    \label{fig:AB}
\end{figure*}

In this work, we integrate the topology-regularized and topology-free patterns for boosting the performance of traffic speed prediction. 
One interesting question may arise: what is the optimal extent of topology-regularized patterns to be accepted for the DCST to achieve the best performance? 
To answer the question, we study the effects of the trade-off parameters $\alpha$ and $\beta$ in Equation~(\ref{equation:kd}). 
The trade-off parameters decide whether the prediction results rely on more topology-free or topology-regularized patterns.

Specifically, we set $\alpha + \beta = 1$, and select five pair of values, {\it i.e.}, $\{(\alpha=0.1, \beta=0.9), (\alpha=0.3, \beta=0.7), (\alpha=0.5, \beta=0.5), (\alpha=0.7, \beta=0.3), (\alpha=0.9, \beta=0.1)\}$, to investigate the corresponding performances. 
The larger $\alpha$ is, the more the prediction relies on the topology-regularized patterns from the GNN-based methods. 
We present the MAPE for each pair of trade-off parameters in Figure~\ref{fig:ex_im}. 

An important observation is that there is not a universally optimal setting for the parameters $\alpha$ and $\beta$ applicable to all Graph Neural Network (GNN)-based methods.
It has been noted that different GNN-based methods attain their peak performance with various combinations of topology-regularized and topology-free patterns. 
This variance can be attributed to the fact that the balance between topology-regularized and topology-free patterns is influenced by the specific characteristics of each GNN-based method. 
These characteristics include distinct architectural designs and diverse strategies for graph construction and learning topology-regularized patterns. 
For instance, STGCN, which is the least effective among the GNN models evaluated, demonstrates an increased need for topology-free patterns to enhance its performance.

\vspace{-1mm}
\subsection{RQ3: Ablation Study of DCST}

In this experiment, we aim to study the necessity of cross-scale consideration, the dynamics of topology-free patterns. 
We construct multiple variants of DCST for the analysis. 
For clarity, we name these variants as
(1) \textbf{w/o ST:} variants that remove Spatial Transformer; 
(2) \textbf{w/o TT:} variants that remove Temporal Transformer; 
(3) \textbf{w/o CS:} variants that only use one specific scale in the Spatial Transformer and Temporal Transformer; 
(4) \textbf{DCST:} a complete version that includes all components.
We present the experimental results on the METRLA dataset in Figure~\ref{fig:AB}. 

We can conclude the following findings: 
(1) The performance of w/o ST decreases the most. A possible explanation is that inter-node spatial interactions contribute more to topology-free patterns compared to intra-node temporal relationships. 
(2) The performance of w/o TT also decreases. Such a result validates the necessity and effectiveness of preserving the dynamics of topology-free patterns. 
(3) Although the effect of w/o CS is better than that of w/o ST and w/o TT, it is still not as good as DCST, which reflects noncomprehensive consideration of cross scales in space and time, leading to incomplete information preservation. 
\vspace{-1mm}
\section{Related Work}

\subsection{Traffic Prediction}
Traffic prediction is initially treated as a time series prediction problem. 
Traditional time series models such as ARMA~\cite{benjamin2003generalized} and ARIMA~\cite{box1970distribution} are not capable of modeling the nonlinear and stochastic features due to the linear nature. 
Then, Graph Neural Networks(GNNs)-besed models are widely used for traffic prediction, this kind of method leverages the topology of the graph to describe the relationships between time series and GNNs to capture correlations.
Earlier GNN-based methods treat traffic networks as a pre-defined graph which aggregates patterns from neighboring connected road segments~\cite{yu2017spatio,li2017diffusion}. 
Then, GWNet\cite{wu2019graph} first constructs a self-adaptive graph through two learnable embedding matrices. 
AGCRN\cite{bai2020adaptive} introduces a node-specific module to construct graph. 
Our work distinguishes itself from these methods by overcoming the limitations of graph topology structure with Dual Cross-Scale Transformer and further integrating the topology-regularized pattern through distillation-style learning framework.

\vspace{-1mm}
\subsection{Knowledge Distillation}
The term ``knowledge distillation'' proposed by\cite{hinton2015distilling} refers to a process in which a well-trained teacher model transfers its knowledge to a student model. 
One crucial role of knowledge distillation is performance enhancement. 
Given the prior knowledge from the teacher models, the student models may have better performance than the teacher models.
Then few samples are illustrated. 
\cite{ahn2019variational} propose a creative framework that develops knowledge transfer by maximizing the information betwixt the teacher network and the student network.
\cite{ahn2019variational} introduced this teacher-student mechanism to the transformer model to deal with image issues.

\vspace{-2mm}
\section{Conclusion Remarks}
In this work, we study the problem of traffic speed prediction. 
The current GNN-based methods exploit topology-regularized patterns with graph topology while neglecting topology-free patterns beyond the graph structure. 
To overcome the limitation, we developed a generic wrapper-style framework to boost current GNN-based methods by integrating topology-free patterns. 
Specifically, we devise a Dual Cross-Scale Transformer architecture with a Spatial Transformer for learning cross-scale topology-free patterns and a Temporal Transformer for capturing the dynamics. 
The topology-regularized patterns are integrated into topology-free patterns with a teacher-student learning framework. 
The proposed framework is flexible and can be applied to any current GNN-based methods without any modification. 
The empirical evaluation validates the necessity of cross-scale topology-free patterns and their dynamics, and the effectiveness of our proposed framework for learning such patterns. 

\section*{Acknowledgments}
This research is funded by the Science and Technology Development Fund (FDCT), Macau SAR (File No. 0123/2023/RIA2, 001/2024/SKL, 0047/2022/A1),
the Natural Science Foundation of China under Grant No. 61836013,
the National Science Foundation (NSF) (Grant No. 2040950, 2006889, 2045567), 
the State Key Laboratory of Internet of Things for Smart City (University of Macau) Open Research Project No. SKL-IOTSC(UM)-2024-2026/ORP/GA02/202,
and the University of Macau (SRG2021-00017-IOTSC, MYRG2022-00048-IOTSC).

\section*{Contribution Statement}
In this work, Yicheng Zhou and Pengfei Wang have an equal contribution, specifically, they led the project, provided theoretical support, and were responsible for the overall model design, code implementation, experimental design and paper writing.
Denghui Zhang, Dingqi Yang, and Yanjie Fu provided guidance in solving complex problems and helped with paper writing.
Pengyang Wang provided valuable feedback on the paper drafts.
All authors reviewed and approved the final manuscript.

\bibliographystyle{named}

\end{document}